\documentclass[10pt,twocolumn,letterpaper]{article}

\usepackage[pagenumbers]{cvpr} 

\usepackage[dvipsnames]{xcolor}

\definecolor{cvprblue}{rgb}{0.21,0.49,0.74}
\usepackage[pagebackref,breaklinks,colorlinks,citecolor=cvprblue]{hyperref}
\usepackage{arydshln}
\usepackage{subcaption} 
\usepackage{booktabs}
\usepackage{multirow}

\usepackage{pifont} 
\usepackage{bbding} 
\def\paperID{12376} 
\def\confName{CVPR}
\def\confYear{2026}

\title{EvalTalker: Learning to Evaluate Real-Portrait-Driven Multi-Subject  \\ Talking Humans}

\author{Yingjie Zhou\textsuperscript{\rm 1,2}~~~~Xilei Zhu\textsuperscript{\rm 1,2}~~~~Siyu Ren\textsuperscript{\rm 2}~~~~Ziyi Zhao\textsuperscript{\rm 2}~~~~Ziwen Wang\textsuperscript{\rm 2}\\Farong Wen\textsuperscript{\rm 1}~~~~Yu Zhou\textsuperscript{\rm 1}~~~~Jiezhang Cao\textsuperscript{\rm 1}~~~~Xiongkuo Min\textsuperscript{\rm 1}~~~~Fengjiao Chen\textsuperscript{\rm 2}\\Xiaoyu Li\textsuperscript{\rm 2}~~~~Xuezhi Cao\textsuperscript{\rm 2}~~~~Guangtao  Zhai\textsuperscript{\rm 1}~~~~Xiaohong Liu\textsuperscript{\rm 1,3} \\
\textsuperscript{\rm 1} Shanghai Jiao Tong University 
\hspace{0.3cm} \textsuperscript{\rm 2} Meituan \hspace{0.3cm} 
\textsuperscript{\rm 3} Shanghai Innovation Institute
}

\begin{document}
\maketitle

\begin{abstract}
Speech-driven Talking Human (TH) generation, commonly known as ``Talker," currently faces limitations in multi-subject driving capabilities. Extending this paradigm to ``Multi-Talker," capable of animating multiple subjects simultaneously, introduces richer interactivity and stronger immersion in audiovisual communication. However, current Multi-Talkers still exhibit noticeable quality degradation caused by technical limitations, resulting in suboptimal user experiences. To address this challenge, we construct \textbf{THQA-MT, the first large-scale \underline{M}ulti-\underline{T}alker-generated \underline{T}alking \underline{H}uman \underline{Q}uality \underline{A}ssessment dataset}, consisting of 5,492 Multi-Talker-generated THs (MTHs) from 15 representative Multi-Talkers using 400 real portraits collected online. Through subjective experiments, we analyze perceptual discrepancies among different Multi-Talkers and identify 12 common types of distortion. Furthermore, we introduce \textbf{EvalTalker, a novel TH quality assessment framework.} This framework possesses the ability to perceive global quality, human characteristics, and identity consistency, while \textbf{integrating Qwen-Sync to perceive multimodal synchrony}. Experimental results demonstrate that EvalTalker achieves superior correlation with subjective scores, providing a robust foundation for future research on high-quality Multi-Talker generation and evaluation.
\vspace{-0.5cm}

\end{abstract}

\begin{figure}[!t]
    \centering
    \vspace{-0.3cm}
    \includegraphics[width =1\linewidth]{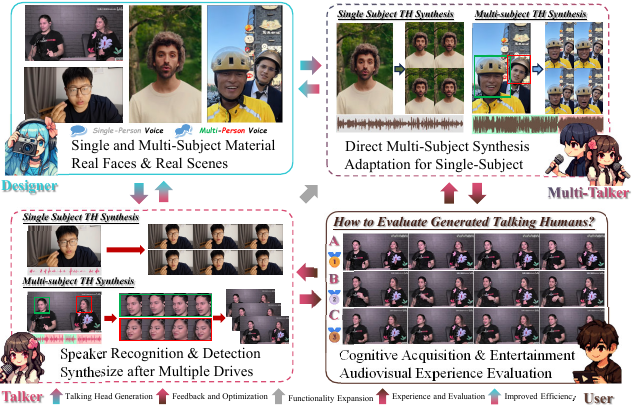}
    \vspace{-0.6cm}
    \caption{Distinction between Talkers and Multi-Talkers. Multi-Talkers extend Talkers with multi-subject driving capabilities, presenting new challenges for quality assessment.}
    \label{fig:teaser}
    \vspace{-0.8cm}
\end{figure}

\begin{table*}[!t]
\centering
\renewcommand\arraystretch{0.9}
\caption{The comparison of digital human quality assessment databases. The ``G" and ``C" denote generative and captive digital human.}
\vspace{-0.2cm}
\resizebox{\linewidth}{!}{\begin{tabular}{cccccccc}
\toprule
Database       & Year  & Modal & Type   & Scale  & Typical Evaluation Methods & Distortion Types& Description            \\
\midrule
DHH-QA \cite{dhhqa} &2023   &Mesh + UV & C & 1,540 & Zhang $et$ $al$. \cite{dhhqa}, Zhou $et$ $al$. \cite{vitqa} & 7 & Scanned Real Human Heads        \\
DDHQA \cite{ddhqa} &2023  &Mesh + UV & C&800    & Zhang $et$ $al$. \cite{zhang2023geometry}, Chen $et$ $al$. \cite{chen2023no}&  9 & Dynamic 3D Digital Human           \\
6G-DTQA \cite{6gqa} &2024 &Mesh + UV& C& 400  & Zhang $et$ $al$. \cite{6gqa} & 5   &   Dynamic 3D Digital Human  \\
THQA-3D \cite{thqa3d}&2024&Mesh + UV& C& 1,000 & Zhou $et$ $al$. \cite{thqa3d}& 5 & Scanned Real Human Heads\\
SJTU-H3D \cite{h3d} &2025 &Mesh + UV & C&1,120  & Zhang $et$ $al$.  \cite{h3d} & 7   & Static 3D Digital Humans  \\ \hdashline
ReLI-QA \cite{reliqa} &2024 & Image & G& 840 & Wen $et$ $al$. \cite{wen2025light} & 4 & Relighted Human Heads \\ 
THQA \cite{thqa} &2024 &Video + Audio& G& 800  & MI3S \cite{zhou2026mi3s}, Xu $et$ $al$. \cite{xu2025facial} & 9   &   AI-Generated Talking Heads  \\
THQA-10K \cite{talker} &2025 &Video + Audio& G & 10,457 & FSCD \cite{talker} & 10 &   AI-Generated Talking Heads\\
THQA-NTIRE \cite{liu2025ntire}&2025 &Video + Audio& C+G & 12,257 &Su $et$ $al$. \cite{su2025quality} & 15 &   2D and 3D Talking Heads\\
AHQA \cite{ahqa}&2025 &Video& G &1,200 & VIP-QA \cite{ahqa} & 4 & Animated Humans\\
MEMO-Bench \cite{zhou2024memo}&2025 & Image & G &7,145 & None & 1 & Emotional Human Heads\\
CDHQA \cite{cdhqa}&2025& Video & C+G &254 & None & 3 &Interactive Digital Human\\
\hline
\bf{THQA-MT (Ours)}  &\bf{2025} &\bf{Video + Audio} & \bf{G} & \bf{5,492}  &\bf{EvalTalker (Ours)} & \bf{12} & \bf{Multi-Subject Talking Humans} \\
\bottomrule
\end{tabular}}
\vspace{-0.5cm}
\label{tab:databases}
\end{table*}
\section{Introduction}
\label{sec:intro}


The rapid advancement of digital media technologies has driven the development of digital humans toward greater realism and anthropomorphism, facilitating their widespread adoption in domains such as live-streaming commerce \cite{chen2024digital}, news broadcasting \cite{kim2022man}, education \cite{guo2024digital}, and cultural dissemination \cite{alfaro2024quality}. Despite these advances, producing lifelike digital humans requires not only strong technical expertise but also substantial time investment, making traditional manual design pipelines inefficient and constraining large-scale applications. In recent years, emergence of Generative Artificial Intelligence (GAI) \cite{sengar2025generative,banh2023generative,ooi2025potential,zhang2024bench,cumt} has provided transformative solutions for digital human creation. Among these, speech-driven talking human generation, commonly referred to as ``Talkers," has become one of the most representative paradigms. By leveraging only speech and portraits, Talkers \cite{audio2head,dreamtalk,iplap,videoretalking,dinet,makelttalk,zhang2025large} can automatically synthesize expressive Talking Human (TH) videos, dramatically simplifying and accelerating the digital human production process. However, as shown in Fig.~\ref{fig:teaser}, most existing Talkers are restricted to animating isolated facial regions of a single subject, overlooking the complex multi-subject dynamics and non-verbal body interactions inherent in real-world human communication. To overcome these limitations, recent works have introduced multi-subject TH generation, known as ``Multi-Talkers," which aim to model natural conversational behaviors among multiple humans \cite{kong2025let,chen2025hunyuanvideo,gan2025omniavatar}. Nonetheless, due to technical constraints, current Multi-Talkers often produce outputs with substantial quality degradations, that severely compromise the user’s audiovisual experience. Consequently, conducting systematic quality assessments of Multi-Talker-generated Talking Humans (MTHs) is crucial, not only for identifying existing limitations and guiding algorithmic improvement but also for enhancing overall quality and user satisfaction.

Despite growing interest in TH generation, existing quality assessment remains confined to single-subject Talkers, neglecting the multi-subject context and the unique perceptual challenges it introduces. To fill this research gap, we establish the first large-scale Multi-Talker-Generated Talking Human Quality Assessment (THQA-MT) dataset. The dataset comprises 5,492 MTHs synthesized from 400 real portraits and their corresponding speeches, covering 15 representative Multi-Talkers to ensure diversity and representativeness. We further conduct subjective experiments with multiple participants to obtain perceptual ratings, which reveal significant quality discrepancies across different Multi-Talkers and highlight shared perceptual challenges. For objective evaluation, we propose EvalTalker, a comprehensive evaluation framework that jointly models global visual quality, human body features, and identity consistency. Furthermore, the proposed Qwen-Sync extends traditional lip-sync evaluation to a generalized multimodal synchronization paradigm, thereby improving EvalTalker's ability to capture holistic perceptual quality. Extensive experiments show that EvalTalker achieves state-of-the-art (SOTA) performance on multiple benchmark datasets. In summary, the main contributions of this paper are as follows:

\begin{itemize} 
\item We construct \textbf{THQA-MT}, the first large-scale dataset for Multi-Talker-generated talking human quality assessment, comprising 5,492 videos synthesized from 15 representative Multi-Talker models. This dataset establishes a new benchmark for evaluating perceptual quality in multi-subject talking human generation.
\item We develop \textbf{Qwen-Sync}, a multimodal synchrony detection module built upon Qwen2.5-Omni \cite{xu2025qwen2}, which extends traditional lip-sync evaluation to a more generalized cross-modal alignment. Compared with SyncNet \cite{chung2017out}, Qwen-Sync demonstrates broader synchrony perception and superior performance on Multi-Talker evaluation.
\item We propose \textbf{EvalTalker}, a comprehensive quality assessment framework that integrates global visual quality, human body features, identity consistency, and multimodal synchrony. EvalTalker achieves SOTA results across five datasets, verifying its robustness and generalizability for both single- and multi-subject talking human evaluation.

\end{itemize}



\begin{figure*}[!t]
    
    \centering
    \includegraphics[width =1\linewidth]{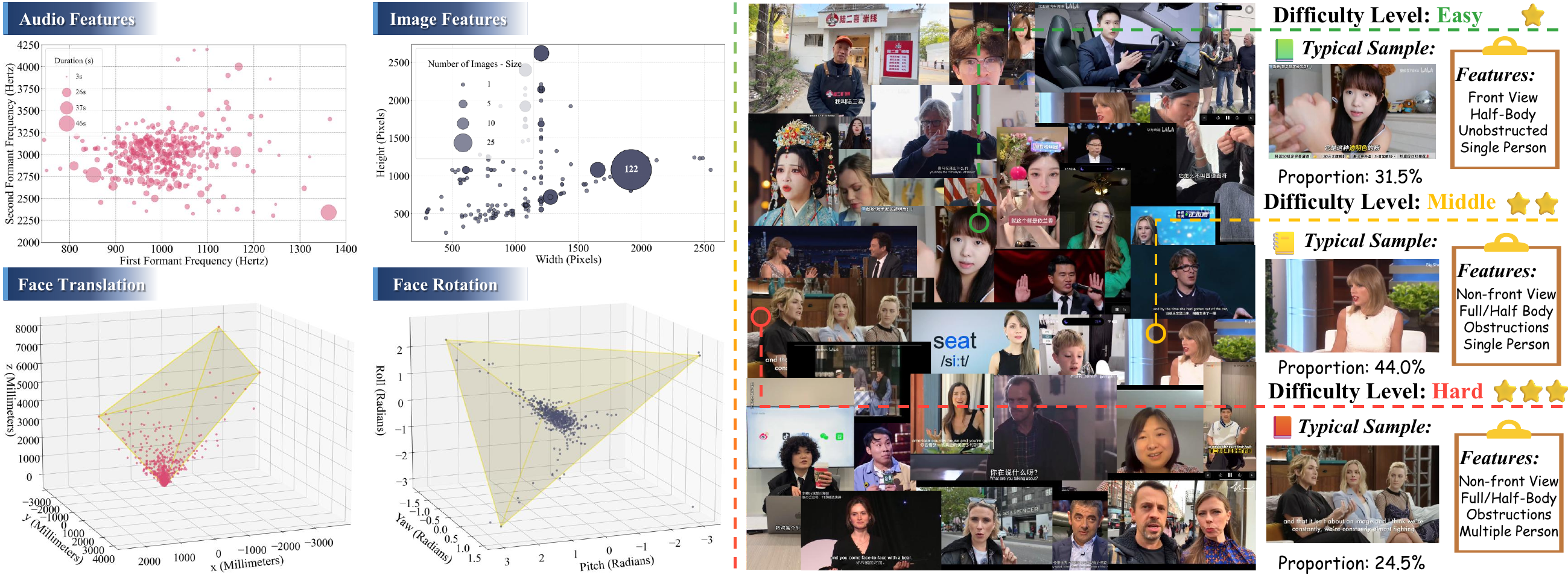}
    \vspace{-0.6cm}
    \caption{Visualization and statistical analysis of selected materials. Images will be anonymized to ensure protection of personal privacy.}
    \label{fig:datasets}
    \vspace{-0.4cm}
\end{figure*}
\section{Related Works}
\label{sec:formatting}

\subsection{Multi-Talker: Multi-Subject Driven Methods}
\label{sec:talker}
Existing speech-driven animation methods are commonly referred to as ``Talkers," while systems capable of animating multiple characters under multi-subject conditions can be classified as ``Multi-Talkers." Currently, Multi-Talkers can be broadly divided into two categories. The first category (\ding{171}) \cite{kong2025let,chen2025hunyuanvideo,gan2025omniavatar} comprises end-to-end speech-driven multi-subject rendering methods, which perform joint inference to animate all subjects simultaneously. These methods automatically detect the number and positions of individuals and can be regarded as true Multi-Talkers in the strictest sense. The second category (\ding{168}) \cite{wav2lip,musetalk,sadtalker,liu2024anitalker,cao2024joyvasa} employs sequential Talker-based pipelines, where each subject is animated independently using traditional Talker, and the outputs are composited via stitching or blending techniques to achieve multi-subject animation effects. Although this strategy enables basic multi-subject synthesis, it suffers from inherent limitations in modeling inter-subject interactions and maintaining spatial and contextual coherence with background environments. Furthermore, some Multi-Talkers extend beyond facial animation by incorporating upper-body or full-body motion generation, while others remain limited to facial region driving. Overall, Multi-Talker represents a significant yet nascent extension of Talker, one that poses new challenges in multimodal coordination, interaction, and perceptual consistency across multiple animated subjects.

\subsection{Digital Human Quality Assessment}
Several representative datasets have been developed to support digital human quality assessment, as summarized in Table~\ref{tab:databases}. Within this domain, Talking Human Quality Assessment (THQA) has emerged as a central research focus. While existing datasets such as THQA-10K \cite{talker} and THQA-NTIRE \cite{liu2025ntire} provide substantial data for THQA research, they are limited in several respects: 1) They primarily focus on the animation of isolated facial regions, neglecting multi-subject interactions and conversational body movements; 2) All 2D talking heads in these datasets are derived from GAI, which differ significantly from real portraits. To address these limitations, we construct THQA-MT dataset by synthesizing MTHs from 400 real portraits paired with corresponding audio. This dataset provides a more realistic and diverse benchmark for MTH evaluation.

Building on these datasets, various targeted quality evaluation methods have been proposed. For example, Su $et$ $al$. \cite{su2025quality} employed a dual-stream network to separately extract video and audio features from THs, enabling multimodal assessment. Xu $et$ $al$. \cite{xu2025facial}, leveraging the Facial Action Coding System (FACS) \cite{ekman1978facial}, captured micro-expressions and facial topology to propose an interpretable THQA method. Zhou $et$ $al$. introduced two evaluation frameworks, FSCD \cite{talker} and MI3S \cite{zhou2026mi3s}, for AI-generated THs: FSCD uses the Y-T Slice \cite{shan2011xt} to capture temporal mouth dynamics, while MI3S integrates image quality, aesthetics, identity consistency, and lip-sync consistency into a comprehensive assessment framework. Despite these advances, existing THQA methods are limited to single-subject speaker faces, largely ignoring natural speech movements and the coherence of multi-subject interactions.

\begin{figure*}[!t]
    
    \centering
    \includegraphics[width =1\linewidth]{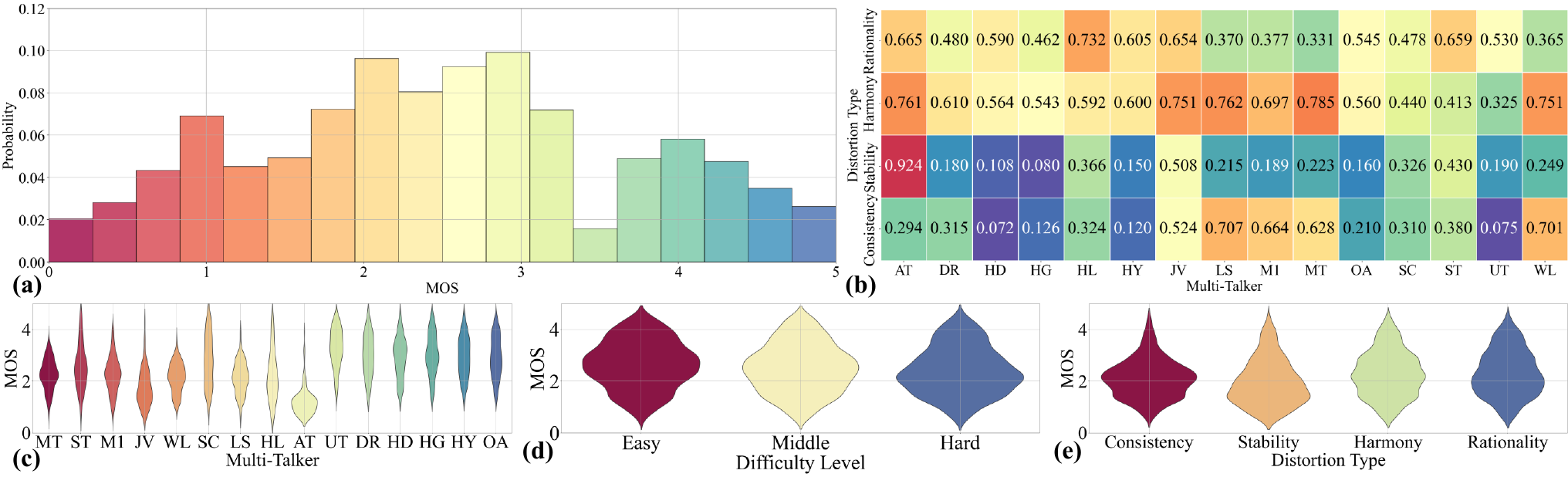}
    \vspace{-0.65cm}
    \caption{Visualization of subjective experimental results. Subfigure (a) shows the MOS distribution of THQA-MT dataset. Subfigure (b) illustrates the distortion rate for Multi-Talkers, while Subfigure (c-e) depict the impact of various factors on the MOS distribution.}
    \label{fig:mos}
    \vspace{-0.45cm}
\end{figure*}

\section{Database Construction}
\subsection{Material Collection}
Unlike existing THQA datasets \cite{thqa,talker,liu2025ntire}, to evaluate Multi-Talkers in real-world scenarios, we collect 400 videos from online sources. For audio processing, we directly extract each video's audio track and, in cases with multiple speakers, merge the relevant tracks. For the visual component, we carefully select a representative source frame from each video to serve as the input image for Multi-Talker synthesis during multi-subject generation. To further characterize the driving difficulty of the source images, we classify the 400 images into three levels: Easy, Medium, and Hard, as illustrated in Fig.~\ref{fig:datasets}. To demonstrate the diversity of the collected dataset, Fig.~\ref{fig:datasets} presents a subset of source images, accompanied by statistical analysis of audio features, source image characteristics, and facial poses. Key observations include: 1) The source images vary in resolution, number of subjects, positions, and backgrounds, providing authentic multi-subject conversational scenarios for Multi-Talkers; 2) Extracted speech exhibits diverse phonetic characteristics. The first formant peaks between 700–1400 Hz, reflecting variability in mouth shapes during articulation, while the second formant peaks between 2250–4250 Hz, indicating differences in tongue positions across samples; 3) Pose estimation by OpenFace \cite{baltruvsaitis2016openface} reveals that the source images encompass a wide range of facial positions and orientations, capturing diverse initial conditions for synthesis.

\begin{table}[!t]
    \centering
    \caption{Details of Multi-Talkers employed. Symbols \ding{171} and \ding{168} are defined in Sec.~\ref{sec:talker}, and $*$ denotes closed-source Multi-Talker. Sample size represents number of successfully generated cases.}
    \vspace{-0.3cm}
    \resizebox{1\linewidth}{!}{\begin{tabular}{c|c|c|c|c|c}
    \toprule
          Label &Methods & Year & Type & Motion & Sample Size\\ \hline
           DR&Dreamina AI$^*$ \cite{jimeng}  & 2025& \ding{171}&\checkmark &400 / 400 \\
        HG&HeyGen$^*$ \cite{heygen} & 2025& \ding{171}&\checkmark &399 / 400 \\
         HD&Hedra$^*$ \cite{hedra} & 2025& \ding{171}&\checkmark &393 / 400  \\ \hdashline
           UT &MultiTalk \cite{kong2025let} &2025 & \ding{171} &\checkmark &400 / 400 \\  
         HY&HunyuanAvatar \cite{chen2025hunyuanvideo} &2025 & \ding{171}&\checkmark &400 / 400 \\ 
          OA&OmniAvatar \cite{gan2025omniavatar} &2025 & \ding{171}&\checkmark &400 / 400 \\ 
          HL& Hallo3 \cite{cui2025hallo3}& 2025& \ding{171}& \checkmark& 341 / 400\\ 
           \hdashline
          WL&Wav2Lip \cite{wav2lip}&2020 & \ding{168}& \ding{53}&397 / 400 \\
          JV& JoyVASA \cite{cao2024joyvasa} & 2024& \ding{168}& \ding{53}&375 / 400\\
          LS& LatentSync \cite{li2024latentsync}& 2024& \ding{168}&\ding{53} &364 / 400\\
          MT&MuseTalk \cite{musetalk}&2024 &\ding{168}& \ding{53} &243 / 400 \\ \hdashline
          ST & SadTalker \cite{sadtalker} &2023& \ding{168} &\checkmark &360 / 400\\
         AT & AniTalker \cite{liu2024anitalker} & 2024& \ding{168} &\checkmark &399 / 400\\
         M1 &  MuseTalk 1.5 \cite{musetalk}& 2025& \ding{168}&\checkmark &249 / 400 \\
         SC& Sonic \cite{ji2025sonic}& 2025& \ding{168}& \checkmark&372 / 400\\

    \bottomrule
    \end{tabular}}
    \label{tab:talker}
    \vspace{-0.4cm}
\end{table}

\subsection{Multi-Talker-Generated Talking Human}
Based on the collected source images and audio materials, we select 15 representative Multi-Talkers for MTH generation. A detailed comparison of their algorithmic characteristics and generative capabilities is presented in Table~\ref{tab:talker}, which provides a comprehensive overview of performance differences across the selected methods. From Table~\ref{tab:talker}, several key observations can be made: 1) The selected Multi-Talkers represent a diverse set of approaches, including both SOTA multi-subject speech-driven algorithms and classical single-subject methods, ensuring comprehensive coverage of existing techniques; 2) In terms of talking motion, certain Multi-Talkers support action or gesture generation alongside facial animation, while others are limited to facial region synthesis, revealing distinct functional capabilities across methods. To further assess the robustness of different Multi-Talkers, we conduct a statistical analysis of the success rate in generating valid MTHs, as summarized in Table~\ref{tab:talker}. The results indicate that most Multi-Talkers maintain strong robustness across diverse inputs, whereas MuseTalk series \cite{musetalk} exhibits stricter input constraints, resulting in a lower rate of successful generation. In total, 15 Multi-Talkers successfully produce 5,492 MTHs, which collectively constitute the THQA-MT dataset.

\subsection{Subjective Experiment}
To obtain authentic user feedback, we conduct a subjective quality assessment experiment involving 40 participants (20 male and 20 female) who evaluate 5,492 MTHs from the constructed THQA-MT dataset. The experiment is carried out in a well-controlled laboratory environment in accordance with the ITU-R BT.500-13 \cite{bt2002methodology} recommendations. All MTHs are displayed on iMac monitors with a native resolution of 4,096 × 2,304, and participants use wireless headphones to ensure low-latency audio playback and prevent potential cross-audio interference among evaluators. The 5,492 MTHs are divided into 28 evaluation sessions, each containing no more than 200 MTHs. To minimize the influence of visual fatigue, participants are required to take a mandatory 30-minute break between sessions. Additionally, each participant can complete a maximum of 3 sessions per day to ensure the reliability of subjective ratings.

\subsection{Data Processing}
In the subjective experiment, we collect 219,680 = 40 × 5,492 subjective evaluations. Following the protocol established for the THQA-10K dataset, each evaluation is represented as a tuple $\{q_{ij}, D_{ij}\}$, where $q_{ij}$ and $D_{ij}$ denote the subjective quality score and distortion type identification for the $j$-th MTH by the $i$-th participant. Specifically, $D_{ij}$ is a 12-dimensional binary distortion vector, with each element corresponding to a distinct distortion type. To normalize individual rating biases, $q_{ij}$ is transformed into a z-score:
\begin{equation}
z_{ij} = \frac{{{q_{ij}} - \mu _i}}{{\sigma _i}},
\end{equation}
where $\mu_{i}=\frac{1}{N_{i}} \sum_{j=1}^{N_{i}} q_{i j}$, $\sigma_{i}=\sqrt{\frac{1}{N_{i}-1} \sum_{j=1}^{N_{i}}\left(q_{i j}-\mu_{i}\right)}$, and $N_i$ represents the total number of MTHs evaluated by subject $i$. In accordance with the rejection procedure described in \cite{bt2002methodology}, ratings from unreliable participants are excluded. The remaining $z_{ij}$ are linearly rescaled to the range [0, 5], and the Mean Opinion Score (MOS) for each MTH is obtained by averaging the rescaled z-scores. For distortion classification, a majority-voting rule is adopted: a distortion type is considered present in the $j$-th MTH only if more than half of the participants identified it as such.

\begin{figure*}[!t]
    
    \centering
    \includegraphics[width =1\linewidth]{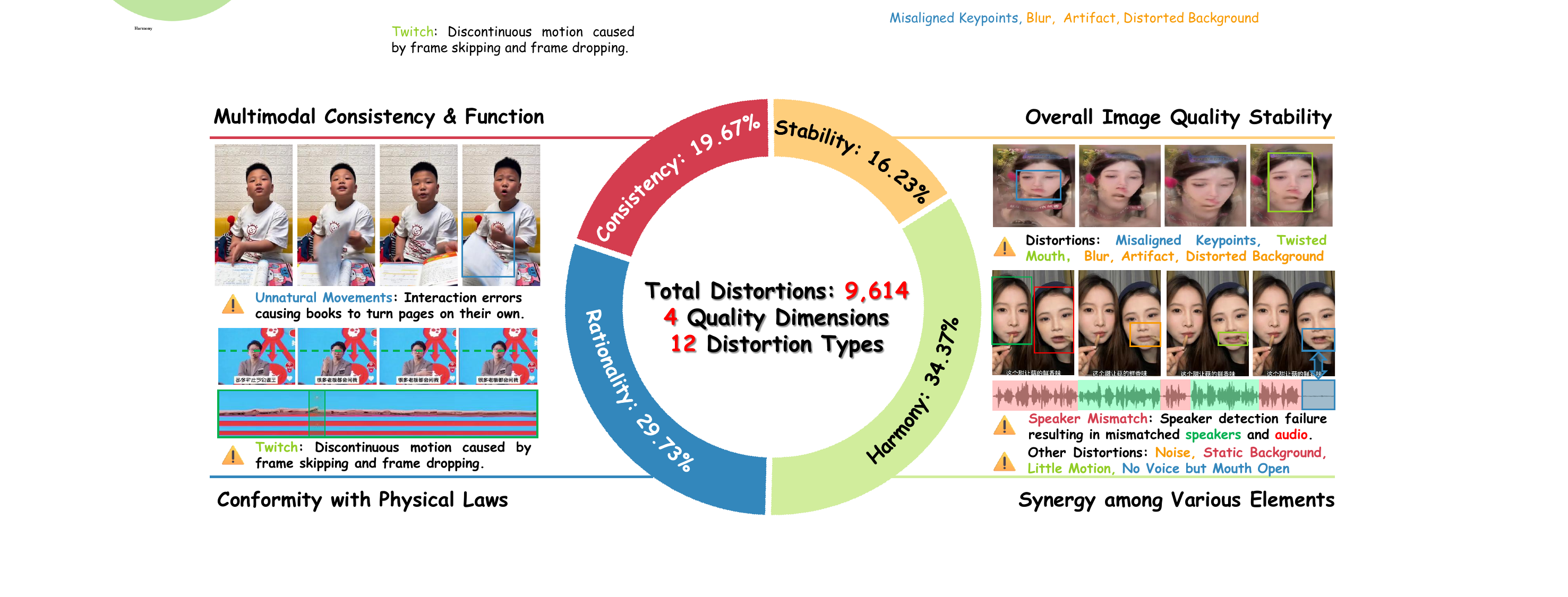}
    \vspace{-0.6cm}
    \caption{Visualization of distortion types and quality dimensions. Colors indicate corresponding quality dimension for each distortion.}
    \label{fig:dis}
    \vspace{-0.4cm}
\end{figure*}
\subsection{Mean Opinion Score Analysis}
To provide an intuitive overview of the MOS distribution of MTHs in the THQA-MT dataset, we present the bar chart in Fig~\ref{fig:mos}(a). To further analyze the influence of multiple factors on MTH quality, Figs.~\ref{fig:mos}(c-e) illustrate the relationships between MOS and specific variables. From these visualizations, several key observations can be made: 1) The MOSs in the THQA-MT dataset exhibit a broad and balanced distribution, with most MTHs falling within the moderate-quality range. This suggests that while current Multi-Talkers achieve acceptable perceptual quality, substantial improvement potential remains. Additionally, the presence of MTHs within both low and high quality ranges highlights the diversity and representativeness of the dataset across different perceptual quality levels; 2) Noticeable variations in MOS distributions are observed across different Multi-Talkers. For instance, MultiTalk \cite{kong2025let} consistently produces higher-quality MTHs, whereas AniTalker \cite{liu2024anitalker} demonstrates weak performance; 3) The complexity of source images exerts a significant influence on Multi-Talker performance. While Multi-Talkers perform well in single-subject driving, their output quality deteriorates as the number of subjects and scene complexity increase. This observation underscores the limitations of current Multi-Talkers in handling multi-subject, real-world scenarios.

\subsection{Distortion Visualization \& Analysis}
To further reveal the quality challenges of MTHs and offer insights for improving Multi-Talkers, we analyze distortion annotations from the subjective experiments and visualize representative cases in Fig.~\ref{fig:dis}. Several observations can be drawn: 1) Across 5,492 MTHs in the THQA-MT dataset, a total of 9,614 distortion instances are recorded, indicating that most MTHs suffer from multiple co-occurring distortions. This highlights the prevalence of degradation in current Multi-Talkers and underscores the necessity of systematic quality assessment; 2) The 12 identified distortion types can be grouped into four broader quality dimensions. Among these, coordination-related distortions are the most prominent, revealing the particular difficulty Multi-Talkers face in achieving coherent multi-subject synchronization; 3) Compared to traditional single-subject THQA, the THQA-MT dataset includes all 10 distortion types reported by Zhou $et$ $al$. \cite{talker} while introducing two additional categories, speaker mismatch and static background, arising from multi-subject and real-world scenarios. Furthermore, the definitions of existing distortion types have been extended beyond facial regions to encompass whole-body quality and environmental consistency.

To examine the relationship between distortions and perceived quality, we further analyze the joint distribution of MOS and distortion annotations, as illustrated in Fig.~\ref{fig:mos}(b) and (e). The following conclusions emerge: 1) Distortion frequency varies significantly across different Multi-Talkers. The MultiTalk \cite{kong2025let} exhibits the lowest distortion rate, consistent with its higher MOS values, while AniTalker \cite{liu2024anitalker} demonstrates frequent distortions and generally lower perceived quality; 2) Among the four quality dimensions, stability-related distortions occur least often but exert the most severe impact on user experience, as they directly affect visual clarity (e.g., blurring or artifacts). In contrast, consistency and rationality distortions appear more frequently but are better tolerated by observers, reflecting differing perceptual sensitivities across distortion types.

\begin{figure*}[!t]
    \centering
    \includegraphics[width =1\linewidth]{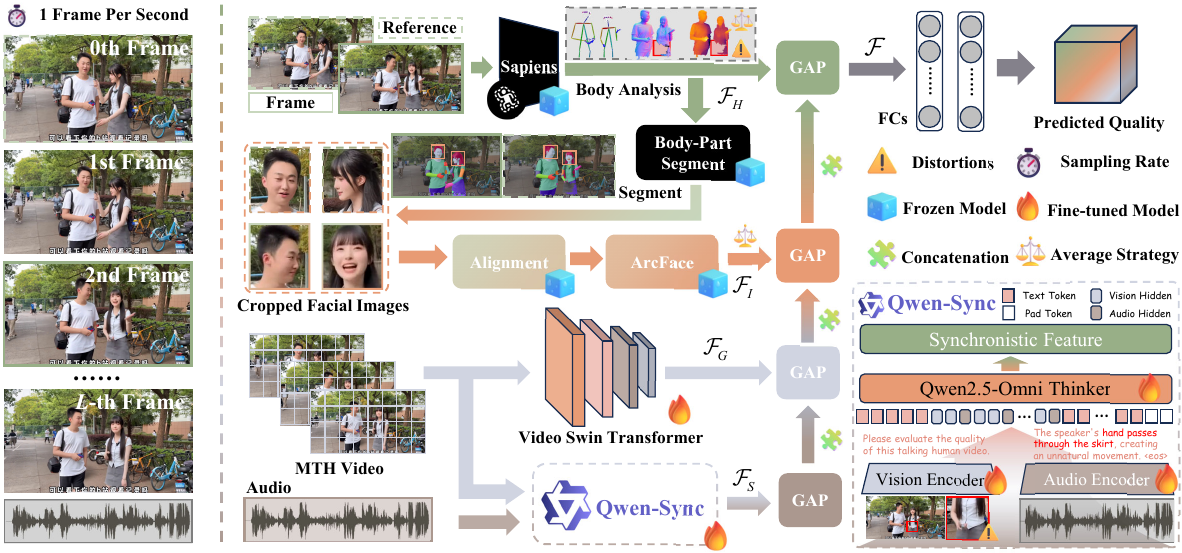}
    \vspace{-0.7cm}
    \caption{The framework of EvalTalker. The framework illustrates a multi-subject case and individuals will be anonymized.}
    \label{fig:evaltalker}
    \vspace{-0.6cm}
\end{figure*}
\section{Proposed Method: EvalTalker}
\subsection{Global Quality Feature Perception}
Based on subjective experimental results indicating that users are particularly sensitive to visual distortions, we first focus on extracting features from the visual perception perspective in EvalTalker. To capture the global quality feature of MTHs, we employ the Video Swin Transformer (VST) \cite{liu2022video}, which effectively models both spatial and temporal dependencies, for feature extraction:
\begin{equation}
    \mathcal{F}_{G} = VST(V),
\end{equation}
where $V$ denotes the MTH video, and $\mathcal{F}_G$ represents global quality features extracted from the MTH via VST backbone.

\begin{table*}[!t]
    \centering
    \setlength{\tabcolsep}{5pt}
    \caption{Performance results on selected five THQA databases and average performance. Best in {\bf\textcolor{red}{RED}}, second in \bf\textcolor{blue}{BLUE}.}
    \vspace{-0.3cm}
    \resizebox{\linewidth}{!}{\begin{tabular}{c|l|cccc|cccc|cccc}
    \toprule
       \multirow{2}{*}{Type} &\multirow{2}{*}{Models}   &  \multicolumn{4}{c|}{THQA} &  \multicolumn{4}{c|}{THQA-3D} &\multicolumn{4}{c}{THQA-10K}\\ \cline{3-14}
       &   & SRCC$\uparrow$ & PLCC$\uparrow$ & KRCC$\uparrow$ & RMSE$\downarrow$ & SRCC$\uparrow$ & PLCC$\uparrow$ & KRCC$\uparrow$ & RMSE$\downarrow$ &SRCC$\uparrow$ & PLCC$\uparrow$ & KRCC$\uparrow$ & RMSE$\downarrow$\\ \hline
       \multirow{4}{*}{IQA} 
        & BRISQUE \cite{brisque}  & 0.4856 & 0.5970& 0.3454&0.8227 & 0.6749& 0.7453&0.5060 &0.5717 & 0.4271& 0.4451&0.2993 &1.0262\\
       & NIQE \cite{niqe} &0.0535	&0.1643	&0.0402	&0.9811 & 0.2243&0.4741 & 0.1232&0.7707 & 0.0089&0.0436 &0.0051 & 1.1492\\
       & CPBD \cite{cpbd} &	0.0575	&0.0876	&0.0376	&0.9908 &0.2145 &0.3136 &0.1432 &0.8273 &0.0553 &0.0686 &0.0371 & 1.1476\\
       & IL-NIQE \cite{ilniqe} & 0.0537	&0.2160	&0.0276	&0.9712 & 0.2293&0.4871 & 0.1537&0.7600& 0.0490& 0.0634& 0.0286& 1.1480\\ \hdashline
       \multirow{2}{*}{Sync} 
       & LSE-C \cite{chung2017out}  &	0.0056	&0.2109	&0.0048	&0.9723 &0.1728 &0.2297 &0.1355 &0.8499& 0.0706& 0.1634& 0.0468& 1.1349\\
       & LSE-D \cite{chung2017out} &	0.1366	&0.2336	&0.0855	&0.9671 & 0.0079 &0.1054 &0.0008 &0.8684& 0.0580& 0.1123& 0.0385& 1.1431\\  \hdashline
       \multirow{10}{*}{VQA} & VIIDEO \cite{viideo} & 0.1777 &0.1891 & 0.1354& 0.9595& 0.1056& 0.2308& 0.0721&  0.8387&0.1354 &0.1782 & 0.0901& 1.1319 \\
       & TLVQM \cite{tlvqm}  & 0.0254 &0.0355& 0.0209&1.0853& 0.1887& 0.3112 &0.1272 &0.8240&0.4377 & 0.4679& 0.3070&1.0130 \\
       & VIDEVAL \cite{videval} & 0.0317 & 0.0358& 0.0231&1.1916&0.2252 & 0.3544& 0.1556& 0.8118 & 0.3869&0.4147 &0.2706 & 1.0431\\
       & V-BLIINDS \cite{vblinds}  & 0.4949 & 0.6403&0.3533&0.7976 &0.5298& 0.6412& 0.3907&0.6674& 0.4740& 0.4977& 0.3334&0.9941\\
       & RAPIQUE \cite{rapique} & 0.1789 & 0.1908& 0.1277&1.0162 & 0.3748& 0.4680& 0.2660&0.7643& 0.3576& 0.3846& 0.2490& 1.0579 \\
       & SimpVQA \cite{simpvqa} & 0.6800 & 0.7592 & 0.5052 & 0.6361&0.6321   & 0.7258&0.4717 &0.5983 & 0.7775& 0.8039& 0.5931& 0.6832\\
       & VSFA \cite{vsfa} &0.7601 & 0.8106& 0.5830&0.5966 & 0.7463&0.7811  &0.5596 &0.5726 & 0.7537& 0.7754& 0.5726&0.7343\\
       & FAST-VQA \cite{fastvqa} & 0.6389 & 0.7441& 0.4677&0.6983 & 0.7778&0.7984 &0.5964 &0.5503& 0.7351& 0.7542& 0.5519& 0.8026\\
      & BVQA \cite{bvqa} & 0.7287 & 0.7985&0.5549 &0.6094 & 0.7871&0.8298 &0.6081 &0.5983&0.6335 & 0.7405& 0.4522& 0.7634 \\ \hdashline
          \multirow{5}{*}{THQA} & MI3S \cite{zhou2026mi3s}& 0.7414&0.8207 & 0.5658 & 0.5683  & 0.7671 & 0.8135 & 0.6090 & 0.4637 & 0.7789 & 0.8106 & 0.6032 & 0.6522\\
          & Xu $et$ $al$. \cite{xu2025facial}& \bf\textcolor{blue}{0.8167} & \bf\textcolor{blue}{0.8669} & \bf\textcolor{blue}{0.6422} & 0.5115& \bf\textcolor{blue}{0.8515}& \bf\textcolor{blue}{0.8617} & \bf\textcolor{blue}{0.6768} & \bf\textcolor{blue}{0.4542} & \bf\textcolor{blue}{0.8267} & \bf\textcolor{blue}{0.8614} & \bf\textcolor{blue}{0.6539} & \bf\textcolor{blue}{0.6070}\\
          & Su $et$ $al$. \cite{su2025quality} & 0.8123 & 0.8580 & 0.6397 & 0.5127&0.7838 & 0.8226 & 0.6371& 0.4660& 0.7951&0.8266 & 0.6176 &0.6477\\
          &FSCD \cite{talker}  &0.7812 & 0.8409 &0.5951 &\bf\textcolor{blue}{0.5055} &0.8235 &0.8505 &0.6463 & 0.4577& 0.8066& 0.8322& 0.6228&0.6333\\ 
          & \textbf{EvalTalker (Ours)} & \textcolor{red}{\textbf{0.8447}} & \textcolor{red}{\textbf{0.9026}} & \textcolor{red}{\textbf{0.6660}} & \textcolor{red}{\textbf{0.4674}} & \textcolor{red}{\textbf{0.8849}} & \textcolor{red}{\textbf{0.9106}} & \textcolor{red}{\textbf{0.6973}} & \textcolor{red}{\textbf{0.4238}} & \textcolor{red}{\textbf{0.8560}} & \textcolor{red}{\textbf{0.8814}} & \textcolor{red}{\textbf{0.6789}} & \textcolor{red}{\textbf{0.5773}}  \\
    \bottomrule
       \toprule
       \multirow{2}{*}{Type} &\multirow{2}{*}{Models}   &  \multicolumn{4}{c|}{THQA-NTIRE} &  \multicolumn{4}{c|}{THQA-MT} &\multicolumn{4}{c}{Average Performance}\\ \cline{3-14}
       &   & SRCC$\uparrow$ & PLCC$\uparrow$ & KRCC$\uparrow$ & RMSE$\downarrow$ & SRCC$\uparrow$ & PLCC$\uparrow$ & KRCC$\uparrow$ & RMSE$\downarrow$ &SRCC$\uparrow$ & PLCC$\uparrow$ & KRCC$\uparrow$ & RMSE$\downarrow$\\ \hline
       \multirow{4}{*}{IQA} 
        & BRISQUE \cite{brisque} & 0.4474 & 0.4842 & 0.3261 & 0.9321 & 0.2051 & 0.2778 & 0.1387 & 1.5398 & 0.4480 & 0.5098 & 0.3231 & 0.9785 \\
       & NIQE \cite{niqe} & 0.0454 & 0.1004 & 0.0343 & 1.0231 & 0.1450 & 0.3639 & 0.0944 & 1.4998 & 0.0954 & 0.2292 & 0.0594 & 1.0848\\
       & CPBD \cite{cpbd} & 0.0535 & 0.0745 & 0.0470 & 1.1525 & 0.1127 & 0.1608 & 0.0731 & 1.1634 & 0.0987 & 0.1410 & 0.0676 & 1.0563\\
       & IL-NIQE \cite{ilniqe} & 0.0598 & 0.1015 & 0.0372 & 1.1097 & 0.1214 & 0.4440 & 0.0731 & 1.6221 & 0.1026 & 0.2624 & 0.0640 & 1.1222\\ \hdashline
       \multirow{2}{*}{Sync} 
       & LSE-C \cite{chung2017out} & 0.0820 & 0.1753 & 0.0666 & 1.0870 & 0.3379 & 0.3887 & 0.2260 & 1.3017 & 0.1337 & 0.2336 & 0.0959 & 1.0691\\
       & LSE-D \cite{chung2017out} & 0.0658 & 0.1349 & 0.0445 & 1.1287 & 0.2912 & 0.2031 & 0.1960 & 1.0157 & 0.1119& 0.1578 & 0.0730 & 1.0246\\  \hdashline
       \multirow{10}{*}{VQA} & VIIDEO \cite{viideo} & 0.1554 & 0.1982 & 0.1091 & 1.1025 & 0.0642 & 0.0691 & 0.0433 & 1.1040 & 0.1276 & 0.1730 & 0.0900 & 1.0273\\
       & TLVQM \cite{tlvqm}  & 0.4026 & 0.4348 & 0.2805 & 1.0280 & 0.3506 & 0.4325 & 0.2462 & 0.9961 & 0.2810 & 0.3363 & 0.1963 & 0.9892\\
       & VIDEVAL \cite{videval} & 0.3530 & 0.3784 & 0.2641 & 1.0418 & 0.4061 & 0.4586 & 0.2849 & 0.9822 & 0.2805 & 0.3283 & 0.1996 & 1.0141\\
       & V-BLIINDS \cite{vblinds} & 0.4860 & 0.5329 & 0.3319 & 0.9329 & 0.5102 & 0.5362 & 0.3792 & 0.9371 & 0.4989 & 0.5696 & 0.3577 & 0.8658\\
       & RAPIQUE \cite{rapique} & 0.3392 & 0.3920 & 0.2482 & 1.0395 & 0.1638 & 0.1793 & 0.1120 & 1.0899 & 0.2828 & 0.3229 & 0.2005 & 0.9935\\
       & SimpVQA \cite{simpvqa} & 0.7508 & 0.7912 & 0.5865 & 0.6740 & 0.5257 & 0.5313 & 0.3730 & 0.9424 & 0.6732 & 0.7222 & 0.5059 & 0.7068\\
       & VSFA \cite{vsfa} & 0.7586 & 0.7794 & 0.5699 & 0.7254 & 0.4548 & 0.4777 & 0.3190 & 0.9861 & 0.6947 & 0.7248 & 0.5208 & 0.7230\\
       & FAST-VQA \cite{fastvqa} & 0.7350 & 0.7449 & 0.5497 & 0.7795 & 0.4926 & 0.5074 & 0.3502& 1.1043 & 0.6758 & 0.7098 & 0.5031 & 0.7870\\
      & BVQA \cite{bvqa}  & 0.6497 & 0.7485 & 0.4712 & 0.7556 & 0.4639 & 0.4908 & 0.3270 & 1.0703 & 0.6525 & 0.7216 & 0.4826 & 0.7594\\ \hdashline
          \multirow{5}{*}{THQA} & MI3S \cite{zhou2026mi3s} & 0.7795 & 0.8205 & 0.6084 & 0.6336 & 0.7509 & 0.7820 & 0.6244 & 0.7233 & 0.7635 & 0.8094 & 0.6021 & 0.6082\\
          & Xu $et$ $al$. \cite{xu2025facial} & \bf\textcolor{blue}{0.8105} & \bf\textcolor{blue}{0.8676} & \bf\textcolor{blue}{0.6419} & \bf\textcolor{blue}{0.5949} & 0.7495 & 0.7664 & 0.6058 & 0.7550 & \bf\textcolor{blue}{0.8149} & \bf\textcolor{blue}{0.8448} & \bf\textcolor{blue}{0.6441} & \bf\textcolor{blue}{0.5845}\\
          & Su $et$ $al$. \cite{su2025quality} & 0.8036 & 0.8453 & 0.6285 & 0.6268 & \bf\textcolor{blue}{0.7798} & 0.7931 & 0.6497 & 0.6991 & 0.7949 & 0.8291 & 0.6345 & 0.5904\\
          &FSCD \cite{talker}  & 0.8102 & 0.8314 & 0.6271 & 0.6018 & 0.7769 & \bf\textcolor{blue}{0.7944} & \bf\textcolor{blue}{0.6564} & \bf\textcolor{blue}{0.6875} & 0.7997 & 0.8298 & 0.6295 & 0.5771\\ 
          & \textbf{EvalTalker (Ours)} & \textcolor{red}{\textbf{0.8720}} & \textcolor{red}{\textbf{0.8981}} & \textcolor{red}{\textbf{0.6837}} & \textcolor{red}{\textbf{0.5402}}& \textcolor{red}{\textbf{0.8674}} & \textcolor{red}{\textbf{0.8893}} & \textcolor{red}{\textbf{0.7135}} & \textcolor{red}{\textbf{0.5937}} & \textcolor{red}{\textbf{0.8652}} & \textcolor{red}{\textbf{0.8964}} & \textcolor{red}{\textbf{0.6878}} & \textcolor{red}{\textbf{0.5204}}\\
    \bottomrule
    \end{tabular}}
    \vspace{-0.7cm}
    \label{tab:performance}
\end{table*}
\subsection{Human Body Feature Extraction}
Existing THQA approaches primarily focus on facial features, neglecting body movements during speech and showing limitations in multi-subject scenarios. To address these gaps, EvalTalker incorporates Sapiens \cite{khirodkar2024sapiens}, a foundational model for human body analysis. Trained on extensive human images, Sapiens is capable of performing a variety of human-oriented tasks, including 2D pose estimation, body segmentation, depth estimation, and surface normal prediction, thereby providing rich information for analyzing the body and behavior of each speaker in MTHs. To reduce computational overhead, MTHs are sampled at one frame per second. For precise human feature extraction, we employ the pre-trained Sapiens-2B encoder as the backbone, performing frame-wise feature extraction:
\begin{equation}
    \mathcal{F}_{H}^{i} = Sapiens(f_i),
\end{equation}
where $\mathcal{F}_H^i$ denotes the human features extracted from the $i$-th sampled frame. Finally, an averaging strategy is applied across all sampled frames to obtain the final feature vector $\mathcal{F}_H$, which characterizes the overall human body representation throughout the entire MTH.
\subsection{Identity Consistency Verification}
Identity fidelity is a critical concern and evaluation metric for GAI, and this also applies to the quality assessment of Multi-Talkers. To enable EvalTalker to capture identity fidelity, the original portrait image is incorporated as the $0$-th frame in the sampled frame sequence. Using Sapiens \cite{khirodkar2024sapiens}, human segmentation is performed on the extracted human features $\mathcal{F}_{H}^{0}$ from the reference portrait. Since identity fidelity is primarily determined by facial characteristics, the face region is localized based on the segmentation results, and a cropped reference face $h_0$ is obtained. In multi-subject MTHs, $h_0^n$ denotes the facial image of the $n$-th subject in the reference frame. For each subsequent sampled frame, the same procedure of body segmentation, face localization, and cropping is applied to obtain $h_i$. Identity consistency between $h_0$ and $h_i$ is computed using the pre-trained Face Alignment \cite{zhang2016joint} and ArcFace algorithm \cite{deng2019arcface}:
\begin{equation}
    {\mathcal{F}_I} = \frac{1}{{NL}}\sum\limits_{n = 1}^N {\sum\limits_{i = 1}^L {Arcf(Align(h_0^n),Align(h_i^n))} },
\end{equation}
where $N$ is the number of subjects in the MTH, $L$ represents the total number of sampled frames, $Align(\cdot)$ and $Arcf(\cdot)$ denote the facial alignment and similarity computation. After averaging across all frames and subjects, ${\mathcal{F}_I}$ represents the average identity consistency between each subject in the MTH video and their corresponding reference portrait.
\subsection{Qwen-Sync} 
In MTHs, multimodal synchrony, including lip-audio, audio-behavior, and audio-emotion alignment, strongly affects perceived quality. Traditional SyncNet \cite{chung2017out} focuses only on lip-audio consistency in single-subject videos, limiting their applicability for multi-subject and broader multimodal evaluation. To address these limitations, we propose Qwen-Sync, a novel coordination detection module built on Qwen-2.5-Omni \cite{xu2025qwen2}. Qwen-2.5-Omni is fully fine-tuned via supervised fine-tuning (SFT) using THQA-MT dataset, leveraging both score and distortion annotations to enhance its capacity for TH evaluation and multimodal synchrony perception. Features extracted from the last hidden state layer of Qwen-Sync are used as multimodal synchrony:
\begin{equation}
    {\mathcal{F}_S} = QwenSync(V,A),
\end{equation}
where $V$ and $A$ are the video and audio tracks of the MTH, and $QwenSync(\cdot)$ represents the processing performed by the synchrony detection module. The resulting synchrony feature ${\mathcal{F}_S}$ effectively characterizes the alignment and consistency of multimodal signals within the MTH.
 
\subsection{Feature Fusion and Regression}
To integrate features from different modalities, we first apply Global Average Pooling (GAP) to 4 feature categories individually and then concatenate the pooled features:
{\small 
\begin{equation}
    \mathcal{F} = GAP({\mathcal{F}_G}) \oplus GAP({\mathcal{F}_H}) \oplus GAP({\mathcal{F}_I}) \oplus GAP({\mathcal{F}_S}),
\end{equation}
} 
where $GAP(\cdot)$ denotes GAP operation, $\oplus$ represents feature concatenation, and $\mathcal{F}$ is the resulting fused quality feature.
The fused feature $\mathcal{F}$ is then passed through two Fully Connected (FC) layers to regress the predicted quality score ${\hat q}$. During training, Mean Squared Error (MSE) is employed as the loss function to enable continuous optimization.


\begin{table*}[!t]
         \caption{Ablation study results on databases, where `\textit{w/o}' stands for `without'. Best in {\bf\textcolor{red}{RED}}, second in {\bf\textcolor{blue}{BLUE}}. }
         \vspace{-0.3cm}
    \resizebox{1\linewidth}{!}{\begin{tabular}{c|c|cccc|cccc|cccc}
    \toprule
       \multirow{2}{*}{Type}& \multirow{2}{*}{Dimension}  &\multicolumn{4}{c|}{THQA} &\multicolumn{4}{c|}{THQA-3D} &\multicolumn{4}{c}{THQA-10K}\\ \cline{3-14}
        & & SRCC$\uparrow$ & PLCC$\uparrow$ & KRCC$\uparrow$ & RMSE$\downarrow$ & SRCC$\uparrow$ & PLCC$\uparrow$ & KRCC$\uparrow$ & RMSE$\downarrow$ & SRCC$\uparrow$ & PLCC$\uparrow$ & KRCC$\uparrow$ & RMSE$\downarrow$\\ \hline
        \multirow{4}{*}{\shortstack{Features\\ Ablation}} &\textit{w/o} $\mathcal{F}_{G}$   & 0.8051 & 0.8519 & 0.6305 & 0.5190 & 0.8233 & 0.8598 & 0.6455 & 0.4538 & 0.7980 & 0.8282 & 0.6222 & 0.6438\\
         & \textit{w/o} $\mathcal{F}_{H}$ & 0.8296 & 0.8762 & 0.6486 & 0.4889 & 0.8657 & 0.8884 & 0.6781& 0.4404 & 0.8371 & 0.8686 & 0.6540 & 0.5845 \\
        & \textit{w/o} $\mathcal{F}_{I}$  & \bf\textcolor{blue}{0.8344} & \bf\textcolor{blue}{0.8907} & \bf\textcolor{blue}{0.6555} & \bf\textcolor{blue}{0.4725} & 0.8737 & 0.8962 & 0.6890 & 0.4371 & \bf\textcolor{blue}{0.8399} & \bf\textcolor{blue}{0.8722} & \bf\textcolor{blue}{0.6663} & \bf\textcolor{blue}{0.5800}\\
        & \textit{w/o} $\mathcal{F}_{S}$   & 0.8119  & 0.8734 & 0.6397 & 0.5050 & 0.8741 & 0.8990 & 0.6886 & 0.4352 & 0.8245 & 0.8575 & 0.6476 & 0.6111\\ \hline
         \multirow{2}{*}{\shortstack{Component \\ Replacement}} & SyncNet & 0.8270 & 0.8871 & 0.6504 & 0.4848 & \bf\textcolor{blue}{0.8777} & \bf\textcolor{blue}{0.9046} & \bf\textcolor{blue}{0.6932} & \bf\textcolor{blue}{0.4295} & 0.8263 & 0.8598 & 0.6504 & 0.6037\\
        & \textbf{Ours}  & \textcolor{red}{\textbf{0.8447}} & \textcolor{red}{\textbf{0.9026}} & \textcolor{red}{\textbf{0.6660}} & \textcolor{red}{\textbf{0.4674}} & \textcolor{red}{\textbf{0.8849}} & \textcolor{red}{\textbf{0.9106}} & \textcolor{red}{\textbf{0.6973}} & \textcolor{red}{\textbf{0.4238}} & \textcolor{red}{\textbf{0.8560}} & \textcolor{red}{\textbf{0.8814}} & \textcolor{red}{\textbf{0.6789}} & \textcolor{red}{\textbf{0.5773}}\\ 
    \bottomrule
    \toprule
      \multirow{2}{*}{Type}& \multirow{2}{*}{Dimension}  &\multicolumn{4}{c|}{THQA-NTIRE} &\multicolumn{4}{c|}{THQA-MT} &\multicolumn{4}{c}{Average Performance}\\ \cline{3-14}
        & & SRCC$\uparrow$ & PLCC$\uparrow$ & KRCC$\uparrow$ & RMSE$\downarrow$ & SRCC$\uparrow$ & PLCC$\uparrow$ & KRCC$\uparrow$ & RMSE$\downarrow$ & SRCC$\uparrow$ & PLCC$\uparrow$ & KRCC$\uparrow$ & RMSE$\downarrow$\\ \hline
        \multirow{4}{*}{\shortstack{Features\\ Ablation}} &\textit{w/o} $\mathcal{F}_{G}$ & 0.8053 & 0.8432 & 0.6264 & 0.6201  & 0.7833 & 0.8035 & 0.6547 & 0.6914 & 0.8030 & 0.8373 & 0.6358 & 0.5856 \\
         & \textit{w/o} $\mathcal{F}_{H}$ & 0.8389 & 0.8707 & 0.6556 & 0.5664  & \bf\textcolor{blue}{0.8460} & \bf\textcolor{blue}{0.8696} & \bf\textcolor{blue}{0.6911} & \bf\textcolor{blue}{0.6158} & 0.8434 & 0.8747 & 0.6654 & 0.5392\\
        & \textit{w/o} $\mathcal{F}_{I}$ & \bf\textcolor{blue}{0.8414} & \bf\textcolor{blue}{0.8770} & \bf\textcolor{blue}{0.6746} & \bf\textcolor{blue}{0.5619} & 0.8355 & 0.8618 & 0.6802 & 0.6323 & \bf\textcolor{blue}{0.8449} & \bf\textcolor{blue}{0.8795} & \bf\textcolor{blue}{0.6731} & \bf\textcolor{blue}{0.5367}\\
        & \textit{w/o} $\mathcal{F}_{S}$ & 0.8248 & 0.8636 & 0.6405 & 0.5877& 0.7998 & 0.8146 & 0.6570 & 0.6777 & 0.8270 & 0.8616 & 0.6546 & 0.5633\\  \hline
         \multirow{2}{*}{\shortstack{Component \\ Replacement}} & SyncNet & 0.8356 & 0.8762 & 0.6593 & 0.5646 & 0.8131 & 0.8344 & 0.6694 & 0.6570 & 0.8359 & 0.8724 & 0.6645 & 0.5479\\
        & \textbf{Ours}  & \textcolor{red}{\textbf{0.8720}} & \textcolor{red}{\textbf{0.8981}} & \textcolor{red}{\textbf{0.6837}} & \textcolor{red}{\textbf{0.5402}}& \textcolor{red}{\textbf{0.8674}} & \textcolor{red}{\textbf{0.8893}} & \textcolor{red}{\textbf{0.7135}} & \textcolor{red}{\textbf{0.5937}} & \textcolor{red}{\textbf{0.8652}} & \textcolor{red}{\textbf{0.8964}} & \textcolor{red}{\textbf{0.6878}} & \textcolor{red}{\textbf{0.5204}} \\ 
    \bottomrule

    \end{tabular}}
    \vspace{-0.6cm}
 \label{tab:abl}
    \end{table*}

\section{Experiments}
\subsection{Experiment Details \& Criteria}
To evaluate the effectiveness of EvalTalker, we conduct comprehensive experiments across 5 THQA datasets using representative objective evaluation methods. For dataset selection, we include existing talking-head-oriented quality assessment datasets, namely: THQA \cite{thqa} and THQA-10K \cite{talker} for 2D talking head evaluation, THQA-3D \cite{thqa3d} for 3D talking head quality-of-experience assessment, and THQA-NTIRE \cite{liu2025ntire}, which encompasses both 2D and 3D scenarios. In addition, the THQA-MT dataset introduced in this work is employed for performance validation. Detailed information on the selected datasets is provided in Table~\ref{tab:databases}. For competitor algorithms, we consider classical Image Quality Assessment (IQA) and Video Quality Assessment (VQA) methods, as well as widely used lip-sync consistency metrics, in addition to existing THQA algorithms.

To ensure the robustness of experimental results, all datasets are partitioned using five-fold cross-validation, guaranteeing no content overlap between folds. Performance is quantified using four standard metrics in objective quality assessment: Spearman’s Rank-Order Correlation Coefficient (SRCC), Pearson’s Linear Correlation Coefficient (PLCC), Kendall’s Rank-Order Correlation Coefficient (KRCC), and Root Mean Square Error (RMSE). The average performance across the five folds is then reported to assess the overall effectiveness of the evaluation method.



\subsection{Performance Analysis}
The evaluation results of various methods on the selected datasets are summarized in Table~\ref{tab:performance}. Several key observations can be made: 1) The proposed EvalTalker achieves SOTA performance across all five datasets, outperforming the second-best method by approximately +5\% SRCC on average, demonstrating its effectiveness in objective quality assessment; 2) EvalTalker attains optimal performance on single-subject talking head quality assessment datasets, indicating that, in addition to evaluating complex multi-subject, real-world scenarios, it is well-suited for assessing the quality of various AI-generated 2D and 3D talking heads, highlighting its robustness, versatility, and generalization capability; 3) From a methodological perspective, THQA methods generally outperform VQA algorithms due to their focus on audio-visual characteristics. EvalTalker further extends traditional lip-audio synchrony evaluation into multimodal consistency, while incorporating human body features tailored for multi-subject scenarios, thereby refining and advancing existing THQA frameworks.


\subsection{Ablation Experiments}
To assess the rationality and effectiveness of each component in EvalTalker, we conduct ablation experiments, including both feature ablation and module replacement. The results are summarized in Table~\ref{tab:abl}, from which several observations can be made: 1) All four feature categories incorporated in EvalTalker contribute positively to overall evaluation performance, confirming the soundness of the four-module design; 2) The relative importance of each feature varies across datasets. For instance, the THQA-3D \cite{thqa3d} dataset primarily emphasizes real-speaker facial quality, exhibiting fewer audio desynchronization and coordination issues, whereas datasets with substantial AI-generated speakers place greater weight on the $\mathcal{F}_s$. Overall, the contribution of each feature aligns with the specific evaluation dimensions emphasized by the respective dataset; 3) Replacing the classic SyncNet \cite{chung2017out} with Qwen-Sync significantly improves performance, demonstrating that Qwen-Sync possesses strong multimodal synchrony perception capabilities.


\section{Conclusion}
Methods capable of achieving multi-subject speech-driven synthesis are vividly termed “Multi-Talkers.” However, due to technical limitations, they inevitably face quality distortion, severely impacting users' experience. To effectively perceive the quality issues present in Multi-Talker-generated Talking Human (MTH) videos, provide valuable guidance for the further development of Multi-Talkers, and enhance the user experience quality, we conduct a comprehensive quality assessment of MTHs. Specifically, we first select 15 Multi-Talkers to synthesize 5,492 MTHs, constructing a large-scale MTH quality assessment (THQA-MT) dataset. Subjective experiments reveal not only significant quality variations among MTHs generated by different Multi-Talkers but also identify 12 common distortion types. Furthermore, we propose EvalTalker by comprehensively considering global quality, human body features, identity consistency, and multimodal synchrony. Experimental results demonstrate that EvalTalker achieves state-of-the-art performance across quality assessment datasets, showing strong alignment with human perception and confirming its effectiveness and generalization capability.


{
    \small
    \bibliographystyle{ieeenat_fullname}
    \bibliography{main}

@String(ECCV= {Eur. Conf. Comput. Vis.})

@String(ECCV  = {ECCV})

@article{cpbd,
  title={A no-reference image blur metric based on the cumulative probability of blur detection (CPBD)},
  author={Narvekar, Niranjan D and Karam, Lina J},
  journal={IEEE Transactions on Image Processing},
  volume={20},
  number={9},
  pages={2678--2683},
  year={2011},
  publisher={IEEE}
}

@article{cumt,
  title={An Implementation of Multimodal Fusion System for Intelligent Digital Human Generation},
  author={Zhou, Yingjie and Chen, Yaodong and Bi, Kaiyue and Xiong, Lian and Liu, Hui},
  journal={arXiv preprint arXiv:2310.20251},
  year={2023}
}

@inproceedings{dhhqa,
  title={Perceptual quality assessment for digital human heads},
  author={Zhang, Zicheng and Zhou, Yingjie and Sun, Wei and Min, Xiongkuo and Wu, Yuzhe and Zhai, Guangtao},
  booktitle={IEEE International Conference on Acoustics, Speech and Signal Processing},
  pages={1--5},
  year={2023}
}

@inproceedings{ddhqa,
  title={Ddh-qa: A dynamic digital humans quality assessment database},
  author={Zhang, Zicheng and Zhou, Yingjie and Sun, Wei and Lu, Wei and Min, Xiongkuo and Wang, Yu and Zhai, Guangtao},
  booktitle={IEEE International Conference on Multimedia and Expo},
  pages={2519--2524},
  year={2023}
}

@article{h3d,
  title={Advancing Zero-Shot Digital Human Quality Assessment through Text-Prompted Evaluation},
  author={Zhang, Zicheng and Sun, Wei and Zhou, Yingjie and Wu, Haoning and Li, Chunyi and Min, Xiongkuo and Liu, Xiaohong and Zhai, Guangtao and Lin, Weisi},
  journal={arXiv preprint arXiv:2307.02808},
  year={2023}
}

@article{6gqa,
  title={Quality-of-Experience Evaluation for Digital Twins in 6G Network Environments},
  author={Zhang, Zicheng and Zhou, Yingjie and Teng, Long and Sun, Wei and Li, Chunyi and Min, Xiongkuo and Zhang, Xiao-Ping and Zhai, Guangtao},
  journal={IEEE Transactions on Broadcasting},
  year={2024},
  publisher={IEEE}
}

@inproceedings{vitqa,
  title={A No-Reference Quality Assessment Method for Digital Human Head},
  author={Zhou, Yingjie and Zhang, Zicheng and Sun, Wei and Min, Xiongkuo and Ma, Xianghe and Zhai, Guangtao},
  booktitle={IEEE International Conference on Image Processing},
  pages={36--40},
  year={2023}
}

@inproceedings{zhang2023geometry,
  title={Geometry-Aware Video Quality Assessment for Dynamic Digital Human},
  author={Zhang, Zicheng and Zhou, Yingjie and Sun, Wei and Min, Xiongkuo and Zhai, Guangtao},
  booktitle={IEEE International Conference on Image Processing},
  pages={1365--1369},
  year={2023}
}

@article{chen2023no,
  title={A no-reference quality assessment metric for dynamic 3D digital human},
  author={Chen, Shi and Zhang, Zicheng and Zhou, Yingjie and Sun, Wei and Min, Xiongkuo},
  journal={Displays},
  volume={80},
  pages={102540},
  year={2023},
  publisher={Elsevier}
}

@inproceedings{sadtalker,
  title={SadTalker: Learning Realistic 3D Motion Coefficients for Stylized Audio-Driven Single Image Talking Face Animation},
  author={Zhang, Wenxuan and Cun, Xiaodong and Wang, Xuan and Zhang, Yong and Shen, Xi and Guo, Yu and Shan, Ying and Wang, Fei},
  booktitle={Proceedings of the IEEE/CVF Conference on Computer Vision and Pattern Recognition},
  pages={8652--8661},
  year={2023}
}

@inproceedings{wav2lip,
  title={A lip sync expert is all you need for speech to lip generation in the wild},
  author={Prajwal, KR and Mukhopadhyay, Rudrabha and Namboodiri, Vinay P and Jawahar, CV},
  booktitle={ACM International Conference on Multimedia},
  pages={484--492},
  year={2020}
}

@article{audio2head,
  title={Audio2head: Audio-driven one-shot talking-head generation with natural head motion},
  author={Wang, Suzhen and Li, Lincheng and Ding, Yu and Fan, Changjie and Yu, Xin},
  journal={arXiv preprint arXiv:2107.09293},
  year={2021}
}

@article{dreamtalk,
  title={DreamTalk: When Expressive Talking Head Generation Meets Diffusion Probabilistic Models},
  author={Ma, Yifeng and Zhang, Shiwei and Wang, Jiayu and Wang, Xiang and Zhang, Yingya and Deng, Zhidong},
  journal={arXiv preprint arXiv:2312.09767},
  year={2023}
}

@InProceedings{iplap,
    author    = {Zhong, Weizhi and Fang, Chaowei and Cai, Yinqi and Wei, Pengxu and Zhao, Gangming and Lin, Liang and Li, Guanbin},
    title     = {Identity-Preserving Talking Face Generation With Landmark and Appearance Priors},
    booktitle = {Proceedings of the IEEE/CVF Conference on Computer Vision and Pattern Recognition},
    month     = {June},
    year      = {2023},
    pages     = {9729-9738}
}

@inproceedings{videoretalking,
  title={Videoretalking: Audio-based lip synchronization for talking head video editing in the wild},
  author={Cheng, Kun and Cun, Xiaodong and Zhang, Yong and Xia, Menghan and Yin, Fei and Zhu, Mingrui and Wang, Xuan and Wang, Jue and Wang, Nannan},
  booktitle={ACM Special Interest Group on Computer Graphics Asia 2022},
  pages={1--9},
  year={2022}
}

@article{dinet,
  title={DINet: Deformation Inpainting Network for Realistic Face Visually Dubbing on High Resolution Video},
  author={Zhang, Zhimeng and Hu, Zhipeng and Deng, Wenjin and Fan, Changjie and Lv, Tangjie and Ding, Yu},
  journal={arXiv preprint arXiv:2303.03988},
  year={2023}
}

@article{makelttalk,
  title={Makelttalk: speaker-aware talking-head animation},
  author={Zhou, Yang and Han, Xintong and Shechtman, Eli and Echevarria, Jose and Kalogerakis, Evangelos and Li, Dingzeyu},
  journal={ACM Transactions On Graphics},
  volume={39},
  number={6},
  pages={1--15},
  year={2020},
  publisher={ACM New York, NY, USA}
}

@article{bt2002methodology,
  title={Methodology for the subjective assessment of the quality of television pictures},
  author={BT, RECOMMENDATION ITU-R},
  journal={International Telecommunication Union},
  year={2002}
}

@article{brisque,
  title={No-reference image quality assessment in the spatial domain},
  author={Mittal, Anish and Moorthy, Anush Krishna and Bovik, Alan Conrad},
  journal={IEEE Transactions on Image Processing},
  volume={21},
  number={12},
  pages={4695--4708},
  year={2012},
  publisher={IEEE}
}

@article{ilniqe,
  title={A feature-enriched completely blind image quality evaluator},
  author={Zhang, Lin and Zhang, Lei and Bovik, Alan C},
  journal={IEEE Transactions on Image Processing},
  volume={24},
  number={8},
  pages={2579--2591},
  year={2015},
  publisher={IEEE}
}

@article{niqe,
  title={Making a “completely blind” image quality analyzer},
  author={Mittal, Anish and Soundararajan, Rajiv and Bovik, Alan C},
  journal={IEEE Signal Processing Letters},
  volume={20},
  number={3},
  pages={209--212},
  year={2012},
  publisher={IEEE}
}

@article{viideo,
  title={A completely blind video integrity oracle},
  author={Mittal, Anish and Saad, Michele A and Bovik, Alan C},
  journal={IEEE Transactions on Image Processing},
  volume={25},
  number={1},
  pages={289--300},
  year={2015},
  publisher={IEEE}
}

@article{vblinds,
  title={Blind prediction of natural video quality},
  author={Saad, Michele A and Bovik, Alan C and Charrier, Christophe},
  journal={IEEE Transactions on Image Processing},
  volume={23},
  number={3},
  pages={1352--1365},
  year={2014},
  publisher={IEEE}
}

@article{tlvqm,
  title={Two-level approach for no-reference consumer video quality assessment},
  author={Korhonen, Jari},
  journal={IEEE Transactions on Image Processing},
  volume={28},
  number={12},
  pages={5923--5938},
  year={2019},
  publisher={IEEE}
}

@article{videval,
  title={UGC-VQA: Benchmarking blind video quality assessment for user generated content},
  author={Tu, Zhengzhong and Wang, Yilin and Birkbeck, Neil and Adsumilli, Balu and Bovik, Alan C},
  journal={IEEE Transactions on Image Processing},
  volume={30},
  pages={4449--4464},
  year={2021},
  publisher={IEEE}
}

@inproceedings{vsfa,
  title={Quality assessment of in-the-wild videos},
  author={Li, Dingquan and Jiang, Tingting and Jiang, Ming},
  booktitle={ACM International Conference on Multimedia},
  pages={2351--2359},
  year={2019}
}

@article{rapique,
  title={RAPIQUE: Rapid and accurate video quality prediction of user generated content},
  author={Tu, Zhengzhong and Yu, Xiangxu and Wang, Yilin and Birkbeck, Neil and Adsumilli, Balu and Bovik, Alan C},
  journal={IEEE Open Journal of Signal Processing},
  volume={2},
  pages={425--440},
  year={2021},
  publisher={IEEE}
}

@inproceedings{simpvqa,
  title={A Deep Learning based No-reference Quality Assessment Model for UGC Videos},
  author={Sun, Wei and Min, Xiongkuo and Lu, Wei and Zhai, Guangtao},
  booktitle={ACM International Conference on Multimedia},
  pages={},
  year={2022}
}

@inproceedings{fastvqa,
  title={Fast-vqa: Efficient end-to-end video quality assessment with fragment sampling},
  author={Wu, Haoning and Chen, Chaofeng and Hou, Jingwen and Liao, Liang and Wang, Annan and Sun, Wenxiu and Yan, Qiong and Lin, Weisi},
  booktitle={European Conference on Computer Vision},
  pages={538--554},
  year={2022},
  publisher={Springer}
}

@article{bvqa,
  title={Blindly Assess Quality of In-the-Wild Videos via Quality-aware Pre-training and Motion Perception},
  author={Li, Bowen and Zhang, Weixia and Tian, Meng and Zhai, Guangtao and Wang, Xianpei},
  journal={IEEE Transactions on Circuits and Systems for Video Technology},
  volume={32},
  number={9},
  pages={5944-5958},
  year={2022}
}

@inproceedings{thqa3d,
  title={Subjective and Objective Quality-of-Experience Assessment for 3D Talking Heads},
  author={Zhou, Yingjie and Zhang, Zicheng and Sun, Wei and Liu, Xiaohong and Min, Xiongkuo and Zhai, Guangtao},
  booktitle={ACM International Conference on Multimedia},
  pages={6033--6042},
  year={2024}
}

@INPROCEEDINGS{thqa,
  author={Zhou, Yingjie and Zhang, Zicheng and Sun, Wei and Liu, Xiaohong and Min, Xiongkuo and Wang, Zhihua and Zhang, Xiao-Ping and Zhai, Guangtao},
  booktitle={IEEE International Conference on Image Processing}, 
  title={Thqa: A Perceptual Quality Assessment Database for Talking Heads}, 
  year={2024},
  volume={},
  number={},
  pages={15-21}}

@article{musetalk,
  title={MuseTalk: Real-Time High Quality Lip Synchronization with Latent Space Inpainting},
  author={Zhang, Yue and Liu, Minhao and Chen, Zhaokang and Wu, Bin and Zeng, Yubin and Zhan, Chao and He, Yingjie and Huang, Junxin and Zhou, Wenjiang},
  journal={arXiv preprint arXiv:2410.10122},
  year={2024}
}

@misc{jimeng,
    title = {{ \url{https://www.dreamina-ai.com/}}},
    author = {Dreamina AI},
    year = {2025}
}

@misc{heygen,
    title = {{ \url{https://www.heygen.com/}}},
    author = {HeyGen},
    year = {2025}
}

@misc{hedra,
    title = {{ \url{https://www.hedra-ai.com/en}}},
    author = {Hedra AI},
    year = {2025}
}

@inproceedings{reliqa,
  title={ReLI-QA: A Multidimensional Quality Assessment Dataset for Relighted Human Heads},
  author={Zhou, Yingjie and Zhang, Zicheng and Wen, Farong and Jia, Jun and Min, Xiongkuo and Wang, Jia  and Zhai, Guangtao},
  booktitle={IEEE Visual Communications and Image Processing},
  year={2024}
}

@inproceedings{chung2017out,
  title={Out of time: automated lip sync in the wild},
  author={Chung, Joon Son and Zisserman, Andrew},
  booktitle={Asian Conference on Computer Vision 2016 Workshops},
  pages={251--263},
  year={2017}
}

@inproceedings{shan2011xt,
  title={An XT slice based method for action recognition},
  author={Shan, Yanhu and Wang, Shiquan and Zhang, Zhang and Huang, Kaiqi},
  booktitle={2011 IEEE International Conference on Computer Vision Workshops},
  pages={1897--1903},
  year={2011}
}

@article{zhang2024bench,
  title={A-Bench: Are LMMs Masters at Evaluating AI-generated Images?},
  author={Zhang, Zicheng and Wu, Haoning and Li, Chunyi and Zhou, Yingjie and Sun, Wei and Min, Xiongkuo and Chen, Zijian and Liu, Xiaohong and Lin, Weisi and Zhai, Guangtao},
  journal={arXiv preprint arXiv:2406.03070},
  year={2024}
}

@article{xu2025facial,
  title={Facial quality assessment of digital humans: A dual-branch framework integrating morphological harmony and expressive coordination},
  author={Xu, Li and Zhou, Yingjie and Liu, Sitong and Wen, Farong and Zhou, Yu and Liu, Xiaohong and Guo, Jie and Wang, Yu and Cao, Jiezhang},
  journal={Displays},
  pages={103221},
  year={2025},
  publisher={Elsevier}
}

@inproceedings{su2025quality,
  title={Quality Assessment for Talking Head Videos via Multi-modal Feature Representation},
  author={Su, Mengjing and Wang, Yi and Chen, Tuo and Li, Chunxiao and Zhao, Shuaiyu and Wen, Jiaxin and Lin, Chuyi and Liu, Sitong and Chu, Ningxin and Zhou, Yu},
  booktitle={Proceedings of the Computer Vision and Pattern Recognition Conference},
  pages={1414--1420},
  year={2025}
}

@article{zhou2026mi3s,
  title={MI3S: A multimodal large language model assisted quality assessment framework for AI-generated talking heads},
  author={Zhou, Yingjie and Zhang, Zicheng and Wu, Sijing and Jia, Jun and Jiang, Yanwei and Sun, Wei and Liu, Xiaohong and Min, Xiongkuo and Zhai, Guangtao},
  journal={Information Processing \& Management},
  volume={63},
  number={1},
  pages={104321},
  year={2025},
  publisher={Elsevier}
}

@article{talker,
  title={Who is a Better Talker: Subjective and Objective Quality Assessment for AI-Generated Talking Heads},
  author={Zhou, Yingjie and Cao, Jiezhang and Zhang, Zicheng and Wen, Farong and Jiang, Yanwei and Jia, Jun and Liu, Xiaohong and Min, Xiongkuo and Zhai, Guangtao},
  journal={arXiv preprint arXiv:2507.23343},
  year={2025}
}

@inproceedings{wen2025light,
  title={A Light-Aware Quality Assessment Method for Relighted Human Heads Based on Multi-Task Learning},
  author={Wen, Farong  and Zhou, Yingjie and Zhang, Zicheng and Liu, Xiaohong and Wang, Jia and Cao, Jiezhang and Wang, Yu and Zhai, Guangtao},
  booktitle={Chinese Conference on Pattern Recognition and Computer Vision (PRCV)},
  pages={0--0},
  year={2025},
  organization={Springer}
}

@article{zhou2024memo,
  title={MEMO-Bench: A Multiple Benchmark for Text-to-Image and Multimodal Large Language Models on Human Emotion Analysis},
  author={Zhou, Yingjie and Zhang, Zicheng and Cao, Jiezhang and Jia, Jun and Jiang, Yanwei and Wen, Farong and Liu, Xiaohong and Min, Xiongkuo and Zhai, Guangtao},
  journal={arXiv preprint arXiv:2411.11235},
  year={2024}
}

@article{ahqa,
  title={Who is a Better Imitator: Subjective and Objective Quality Assessment of Animated Humans},
  author={Zhou, Yingjie and Zhang, Zicheng and Jia, Jun and Jiang, Yanwei and Liu, Xiaohong and Min, Xiongkuo and Zhai, Guangtao},
  journal={IEEE Transactions on Circuits and Systems for Video Technology},
  year={2025},
  publisher={IEEE}
}

@inproceedings{liu2025ntire,
  title={NTIRE 2025 XGC Quality Assessment Challenge: Methods and Results},
  author={Liu, Xiaohong and Min, Xiongkuo and Hu, Qiang and Zhang, Xiaoyun and Guo, Jie and Zhai, Guangtao and Wang, Shushi and Zhou, Yingjie and Liu, Lu and Li, Jingxin and others},
  booktitle={Proceedings of the Computer Vision and Pattern Recognition Conference},
  pages={1389--1402},
  year={2025}
}

@inproceedings{cdhqa,
  title={CDHQA: A Quality Assessment Database for Conversational Digital Human},
  author={Zhou, Yingjie and Wan, Jing and Liu, Sitong and Xia, Yinghan and Lu, Zhixiang and Wen, Farong and Zhang, Zicheng and Wang, Yu and Zhou, Yu and Liu, Xiaohong and others},
  booktitle={International Conference on Image and Graphics},
  pages={15--26},
  year={2025}
}

@article{alfaro2024quality,
  title={Quality of life in the urban context, within the paradigm of digital human capital},
  author={Alfaro-Navarro, Jos{\'e}-Luis and Lopez-Ruiz, V{\'\i}ctor-Ra{\'u}l and Huete-Alcocer, Nuria and Nevado-Pena, Domingo},
  journal={Cities},
  volume={153},
  pages={105284},
  year={2024},
  publisher={Elsevier}
}

@article{chen2024digital,
  title={Digital human technology in the application of live streaming in social media},
  author={Chen, Xi and Ramasamy, Siva Shankar and She, Bibi},
  journal={Radioelectronic and Computer Systems},
  volume={2024},
  number={4},
  pages={34--45},
  year={2024}
}

@article{kim2022man,
  title={Man vs. machine: Human responses to an AI newscaster and the role of social presence},
  author={Kim, Jihyun and Xu, Kun and Merrill Jr, Kelly},
  journal={The Social Science Journal},
  pages={1--13},
  year={2022},
  publisher={Taylor \& Francis}
}

@inproceedings{guo2024digital,
  title={Digital Human Techniques for Education Reform},
  author={Guo, Peirong and Zhang, Qi and Tian, Chunwei and Xue, Wanli and Feng, Xiaocheng},
  booktitle={Proceedings of the 2024 7th International Conference on Educational Technology Management},
  pages={173--178},
  year={2024}
}

@article{sengar2025generative,
  title={Generative artificial intelligence: a systematic review and applications},
  author={Sengar, Sandeep Singh and Hasan, Affan Bin and Kumar, Sanjay and Carroll, Fiona},
  journal={Multimedia Tools and Applications},
  volume={84},
  number={21},
  pages={23661--23700},
  year={2025},
  publisher={Springer}
}

@article{banh2023generative,
  title={Generative artificial intelligence},
  author={Banh, Leonardo and Strobel, Gero},
  journal={Electronic Markets},
  volume={33},
  number={1},
  pages={63},
  year={2023},
  publisher={Springer}
}

@article{ooi2025potential,
  title={The potential of generative artificial intelligence across disciplines: Perspectives and future directions},
  author={Ooi, Keng-Boon and Tan, Garry Wei-Han and Al-Emran, Mostafa and Al-Sharafi, Mohammed A and Capatina, Alexandru and Chakraborty, Amrita and Dwivedi, Yogesh K and Huang, Tzu-Ling and Kar, Arpan Kumar and Lee, Voon-Hsien and others},
  journal={Journal of Computer Information Systems},
  volume={65},
  number={1},
  pages={76--107},
  year={2025},
  publisher={Taylor \& Francis}
}

@inproceedings{cui2025hallo3,
  title={Hallo3: Highly dynamic and realistic portrait image animation with video diffusion transformer},
  author={Cui, Jiahao and Li, Hui and Zhan, Yun and Shang, Hanlin and Cheng, Kaihui and Ma, Yuqi and Mu, Shan and Zhou, Hang and Wang, Jingdong and Zhu, Siyu},
  booktitle={Proceedings of the Computer Vision and Pattern Recognition Conference},
  pages={21086--21095},
  year={2025}
}

@article{li2024latentsync,
  title={LatentSync: Taming Audio-Conditioned Latent Diffusion Models for Lip Sync with SyncNet Supervision},
  author={Li, Chunyu and Zhang, Chao and Xu, Weikai and Lin, Jingyu and Xie, Jinghui and Feng, Weiguo and Peng, Bingyue and Chen, Cunjian and Xing, Weiwei},
  journal={arXiv preprint arXiv:2412.09262},
  year={2024}
}

@article{cao2024joyvasa,
  title={JoyVASA: portrait and animal image animation with diffusion-based audio-driven facial dynamics and head motion generation},
  author={Cao, Xuyang and Wang, Guoxin and Shi, Sheng and Zhao, Jun and Yao, Yang and Fei, Jintao and Gao, Minyu},
  journal={arXiv preprint arXiv:2411.09209},
  year={2024}
}

@article{gan2025omniavatar,
  title={OmniAvatar: Efficient Audio-Driven Avatar Video Generation with Adaptive Body Animation},
  author={Gan, Qijun and Yang, Ruizi and Zhu, Jianke and Xue, Shaofei and Hoi, Steven},
  journal={arXiv preprint arXiv:2506.18866},
  year={2025}
}

@article{chen2025hunyuanvideo,
  title={HunyuanVideo-Avatar: High-Fidelity Audio-Driven Human Animation for Multiple Characters},
  author={Chen, Yi and Liang, Sen and Zhou, Zixiang and Huang, Ziyao and Ma, Yifeng and Tang, Junshu and Lin, Qin and Zhou, Yuan and Lu, Qinglin},
  journal={arXiv preprint arXiv:2505.20156},
  year={2025}
}

@inproceedings{ji2025sonic,
  title={Sonic: Shifting focus to global audio perception in portrait animation},
  author={Ji, Xiaozhong and Hu, Xiaobin and Xu, Zhihong and Zhu, Junwei and Lin, Chuming and He, Qingdong and Zhang, Jiangning and Luo, Donghao and Chen, Yi and Lin, Qin and others},
  booktitle={Proceedings of the Computer Vision and Pattern Recognition Conference},
  pages={193--203},
  year={2025}
}

@inproceedings{liu2024anitalker,
  title={Anitalker: animate vivid and diverse talking faces through identity-decoupled facial motion encoding},
  author={Liu, Tao and Chen, Feilong and Fan, Shuai and Du, Chenpeng and Chen, Qi and Chen, Xie and Yu, Kai},
  booktitle={Proceedings of the 32nd ACM International Conference on Multimedia},
  pages={6696--6705},
  year={2024}
}

@article{kong2025let,
  title={Let Them Talk: Audio-Driven Multi-Person Conversational Video Generation},
  author={Kong, Zhe and Gao, Feng and Zhang, Yong and Kang, Zhuoliang and Wei, Xiaoming and Cai, Xunliang and Chen, Guanying and Luo, Wenhan},
  journal={arXiv preprint arXiv:2505.22647},
  year={2025}
}

@article{xu2025qwen2,
  title={Qwen2. 5-omni technical report},
  author={Xu, Jin and Guo, Zhifang and He, Jinzheng and Hu, Hangrui and He, Ting and Bai, Shuai and Chen, Keqin and Wang, Jialin and Fan, Yang and Dang, Kai and others},
  journal={arXiv preprint arXiv:2503.20215},
  year={2025}
}

@article{ekman1978facial,
  title={Facial action coding system},
  author={Ekman, Paul and Friesen, Wallace V},
  journal={Environmental Psychology \& Nonverbal Behavior},
  year={1978}
}

@inproceedings{baltruvsaitis2016openface,
  title={Openface: an open source facial behavior analysis toolkit},
  author={Baltru{\v{s}}aitis, Tadas and Robinson, Peter and Morency, Louis-Philippe},
  booktitle={2016 IEEE winter conference on applications of computer vision (WACV)},
  pages={1--10},
  year={2016},
  organization={IEEE}
}

@inproceedings{liu2022video,
  title={Video swin transformer},
  author={Liu, Ze and Ning, Jia and Cao, Yue and Wei, Yixuan and Zhang, Zheng and Lin, Stephen and Hu, Han},
  booktitle={Proceedings of the IEEE/CVF conference on computer vision and pattern recognition},
  pages={3202--3211},
  year={2022}
}

@inproceedings{khirodkar2024sapiens,
  title={Sapiens: Foundation for human vision models},
  author={Khirodkar, Rawal and Bagautdinov, Timur and Martinez, Julieta and Zhaoen, Su and James, Austin and Selednik, Peter and Anderson, Stuart and Saito, Shunsuke},
  booktitle={ECCV},
  pages={206--228},
  year={2024},
  organization={Springer}
}

@inproceedings{deng2019arcface,
  title={Arcface: Additive angular margin loss for deep face recognition},
  author={Deng, Jiankang and Guo, Jia and Xue, Niannan and Zafeiriou, Stefanos},
  booktitle={Proceedings of the IEEE/CVF conference on computer vision and pattern recognition},
  pages={4690--4699},
  year={2019}
}

@article{zhang2016joint,
  title={Joint face detection and alignment using multitask cascaded convolutional networks},
  author={Zhang, Kaipeng and Zhang, Zhanpeng and Li, Zhifeng and Qiao, Yu},
  journal={IEEE signal processing letters},
  volume={23},
  number={10},
  pages={1499--1503},
  year={2016},
  publisher={IEEE}
}

@article{zhang2025large,
  author    = {Zhang, Zicheng and Wang, Junying and Wen, Farong and Guo, Yijin and Zhao, Xiangyu and Fang, Xinyu and Ding, Shengyuan and Jia, Ziheng and Xiao, Jiahao and Shen, Ye and Zheng, Yushuo and Zhu, Xiaorong and Wu, Yalun and Jiao, Ziheng and Sun, Wei and Chen, Zijian and Zhang, Kaiwei and Fu, Kang and Cao, Yuqin and Hu, Ming and Zhou, Yue and Zhou, Xuemei and Cao, Juntai and Zhou, Wei and Cao, Jinyu and Li, Ronghui and Zhou, Donghao and Tian, Yuan and Zhu, Xiangyang and Li, Chunyi and Wu, Haoning and Liu, Xiaohong and He, Junjun and Zhou, Yu and Liu, Hui and Zhang, Lin and Wang, Zesheng and Duan, Huiyu and Zhou, Yingjie and Min, Xiongkuo and Jia, Qi and Zhou, Dongzhan and Zhang, Wenlong and Cao, Jiezhang and Yang, Xue and Yu, Junzhi and Zhang, Songyang and Duan, Haodong and Zhai, Guangtao},
  title     = {Large Multimodal Models Evaluation: A Survey},
  journal   = {SCIENCE CHINA Information Sciences},
  year      = {2025},
  number  = {12},
  volume    = {68},
  pages     = {221301-221369},
  doi       = {https://doi.org/10.1007/s11432-025-4676-4}
}
}


\end{document}